\documentclass[letterpaper, 10 pt, conference]{ieeeconf}  

\IEEEoverridecommandlockouts                              

\usepackage{graphicx}
\usepackage{algorithm}
\usepackage{tabularx}
\usepackage{amsmath, amssymb}
\usepackage{subcaption}
\usepackage{cite}

\usepackage{algpseudocode}

\title{\LARGE \bf
Bridging the Gap: Regularized Reinforcement Learning for Improved Classical Motion Planning with Safety Modules
}

\author{Elias~Goldsztejn$^{1}$, 
        Ronen I. Brafman$^{2}$
\thanks{*This work was supported by ISF Grant 1651/19, the Helmsley Charitable Trust through the ABC Robotics Center of Ben-Gurion University, and the Lynn and William Frankel Center for Computer Science.}
\thanks{$^{1}$Elias Goldsztejn
and $^{2}$Ronen Brafman are with the Department of Computer Science at
        Ben-Gurion University of the Negev.
        {\tt\small eliasgol@post.bgu.ac.il, brafman@bgu.ac.il}}%
}

\begin{document}

\maketitle
\thispagestyle{empty}
\pagestyle{empty}

\begin{abstract}
Classical navigation planners 
can provide safe navigation, albeit often suboptimally and with hindered human norm compliance. 
ML-based, contemporary autonomous navigation algorithms
can imitate more natural and human-compliant navigation, but
usually require large and realistic datasets and do not always provide safety guarantees.
We present an approach that leverages a classical algorithm to guide reinforcement learning. This greatly improves the results and convergence rate of the underlying RL algorithm and requires no
human-expert demonstrations to jump-start the process.
Additionally, we incorporate a practical fallback system that can switch back to a classical planner to ensure safety. 
The outcome is a sample efficient ML approach for mobile navigation that builds on classical algorithms, improves them to ensure human compliance, and guarantees safety.

\end{abstract}

\section{INTRODUCTION}

 The proliferation of self-driving cars by an increasing number of companies and the ability of robots to efficiently deliver food and supplies are clear manifestations of the vast improvements in autonomous navigation.
Nonetheless, various unresolved challenges persist across diverse domains. 

Classical planners use analytic optimization techniques and reactive rules for collision avoidance and for finding safe 
paths~\cite{DWA,ORCA,etb}. These methods can be successful in specific domains, require no or little data, are well understood and can lead to safe and interpretable behavior. However, they often exhibit unnatural and inefficient behaviors and poor social norm compliance~\cite{challenges_social}.

Machine learning methodologies build on the latest advancements in imitation learning (e.g.,
\cite{agile, conditional}) and deep reinforcement learning (e.g.,
\cite{LGM2017,MDD2016,autorl,sacadrl,evita}) through which they
can capture the intricacy of human actions and provide enhanced environmental awareness and human-aware behavior. 
However, they require large and realistic datasets, and often lack safety guarantees~\cite{Survey_1}. 
 Furthermore, because they are often based on end-to-end learning, they lack interpretability and transparency~\cite{interpretability}. For these reasons, they can fail badly in unexpected ways on particular inputs 
\cite{intriguing,evaluation}, making it difficult to rely on them.

This work seeks to alleviate the shortcomings of classical and learning-based methods by combining suitable components of each, building on and modifying various existing methodologies, reviewed in the next section.  In particular, we exploit classical algorithms to improve  the sample efficiency of a learning algorithm and the performance and safety of its resulting navigation policy. 

\begin{figure}[t]
    \centering
    \includegraphics[width=0.25\textwidth]{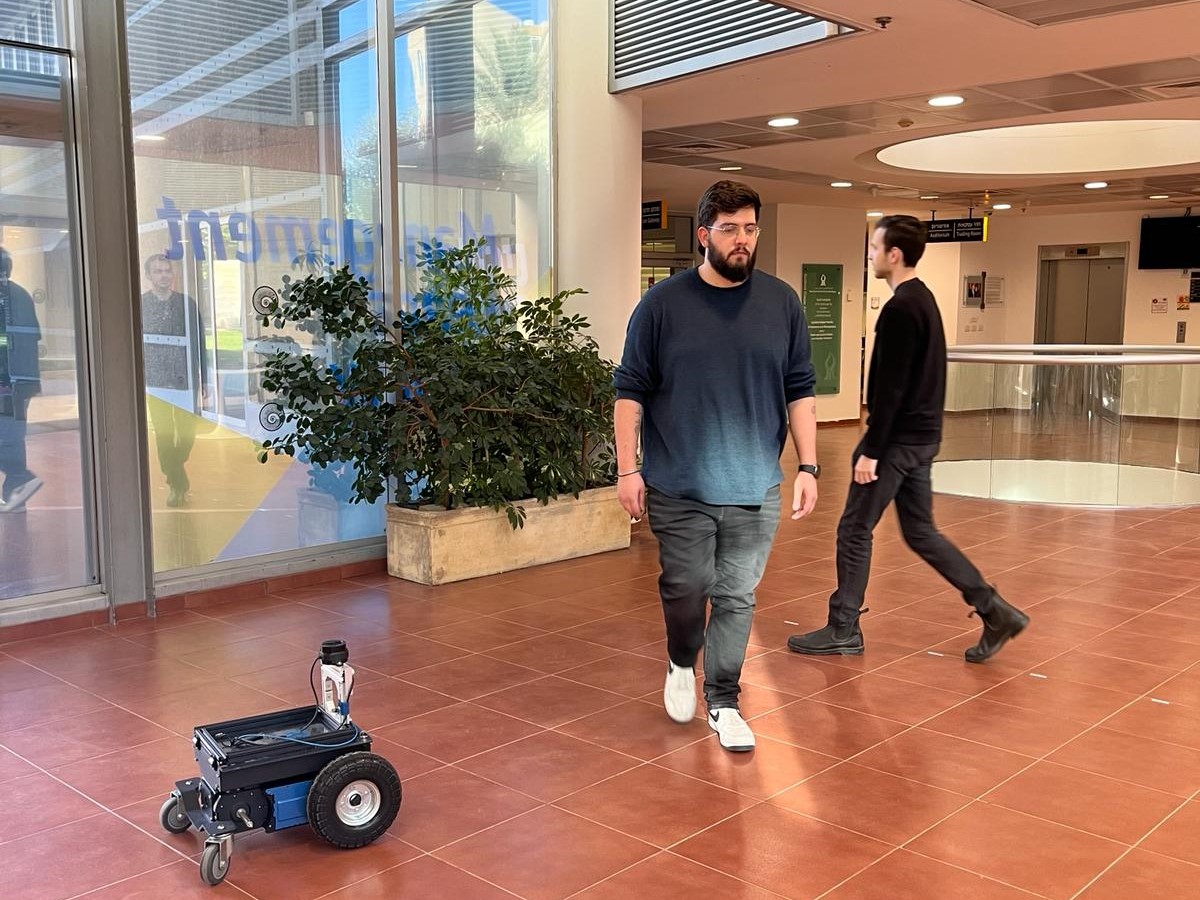}
    \caption{ A robot navigating autonomously 
    using a system that integrates an RL-based policy, regularized with a classical planner, alongside a safety switching mechanism. 
    }
    \label{fig:Physical}
\end{figure}

Our main contribution is a sample-efficient learning strategy for improving classical planners and a fallback system with a trained supervisor that guarantees safety. More specifically,
we suggest the following approach:
\begin{enumerate} 
    \item Train a planner using DRL with policy guidance derived from a classical planner: We seed 
    the replay buffer with experiences generated by the classical planner and 
    regularize the actor in an actor-critic algorithm using the classical planner's policy.
    \item Use a classical rule-based navigation policy as a fallback system and train a supervisor that performs minimal switching between the neural and classical planner to ensure safety.
\end{enumerate}
While our focus is navigation planning, the above offers a general recipe for using learning to improve classical algorithms while retaining 
their respective benefits.
Our approach provides safety and offers transparency at the supervisor level. Its reliance on good, existing classical algorithms helps jump-start the learning algorithm, leading to faster and better convergence, as our empirical evaluation clearly demonstrates. The regularization term further stabilizes the learning process and ensures greater transparency, forcing it to remain in the vicinity of the well-understood classical algorithm. And unlike methods that rely on human demonstrations to achieve some of these effects, no human involvement is needed -- as it is already implicitly present in the formulation of the classical algorithm and the reward 
function.

Relevant code and videos for this paper are available at {\tt\small https://github.com/eliasgoldsztejn95}

\begin{figure}[t]
    \centering
    \includegraphics[width=0.35\textwidth]{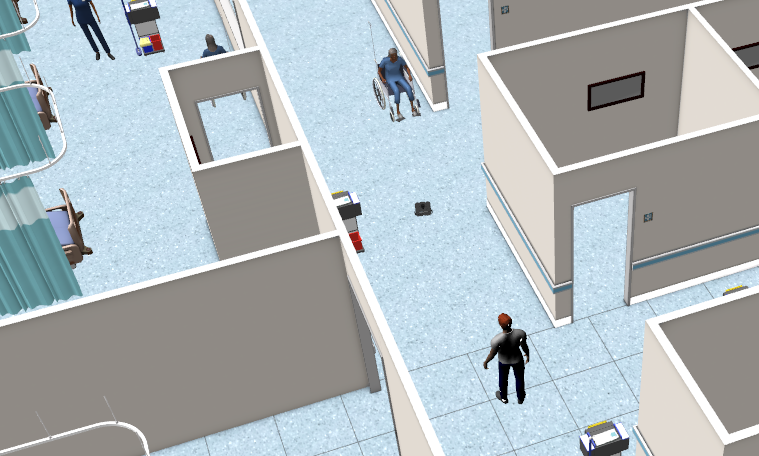}
    \caption{All Learning is
    carried out in this realistic simulated hospital with people moving according to social forces.}
    \label{fig:robot_2}
\end{figure}

\section{Related Work}

We review previous work on classical planners for autonomous driving in mobile robots, reinforcement learning, especially 
using learning from demonstrations, as well as techniques for safe planning.

\subsection{Classical Planning for Mobile Robots}

Early planning algorithms for mobile robots used reactive rules and analytic optimization to avoid collision and reach targets.
Important approaches include Gaussian processes \cite{gaussian}, decentralized and scalable multi-agent reciprocal optimizations \cite{ORCA}, reactive techniques \cite{DWA}, spatial and temporal constraint-aware elastic-band methods \cite{etb}, and more.
Classical planners serve as foundational benchmarks in mobile navigation and are commonly integrated into critical robotic systems, exemplified by their inclusion in the widely adopted ROS \cite{ROS} navigation stack. Although in some environments, these methods can provide effective navigation, they usually require fine-tuning from experts and are limited by their lack of human compliance and social norms, non-smooth behavior, and freezing robot problems, among others.

\subsection{Imitation and Reinforcement Learning}

Imitation learning (IL) and reinforcement learning (RL) are effective and popular techniques for autonomous driving that
can create flexible, human-compliant navigation systems. 
IL uses supervised learning to map states to actions so as to mimic the behavior of a demonstrator, such as a human expert. Some pioneering works include using synthesized data in the form of perturbations \cite{chauffeurnet}, and applying command-conditions to represent expert's intentions \cite{contionionalIL}.
IL has well-known limitations, such as dataset bias and overfitting, new generalization issues, covariate shifts, and requires large amounts of data \cite{ILlimitations}. Mitigating these effects is an ongoing area of interest. Proposed methods include utilizing space-time cost volumes \cite{costvolumes}, which enable interpretability and faster learning. Others include combining reinforcement learning to address the covariate shift problem \cite{imitationnotenough}.

In RL, agents learn to make decisions by interacting with an environment. Feedback is obtained in the form of rewards from state-action pairs, and the primary objective is to derive a policy that maximizes the cumulative reward over time. In contrast with IL, RL has a closed-loop exploration-exploitation approach, which addresses the stability and overfitting challenges of IL.

RL is used extensively for autonomous driving and navigation
\cite{LGM2017,MDD2016,autorl,sacadrl,evita},
lane following and urban driving \cite{lane, chen2021interpretable}. While RL is a promising approach
it faces numerous challenges,
including the difficulty of defining a suitable reward function, the need for extensive training datasets, and concerns related to safety and interpretability \cite{kiran2021deep, xiao2022motion}.
Techniques like inverse and safe RL and RL with demonstrations and human feedback have been developed aimed at addressing these challenges more effectively. However, in many domains, solutions for these issues are still elusive.

\begin{figure*}
    \centering
    \includegraphics[width=7in]{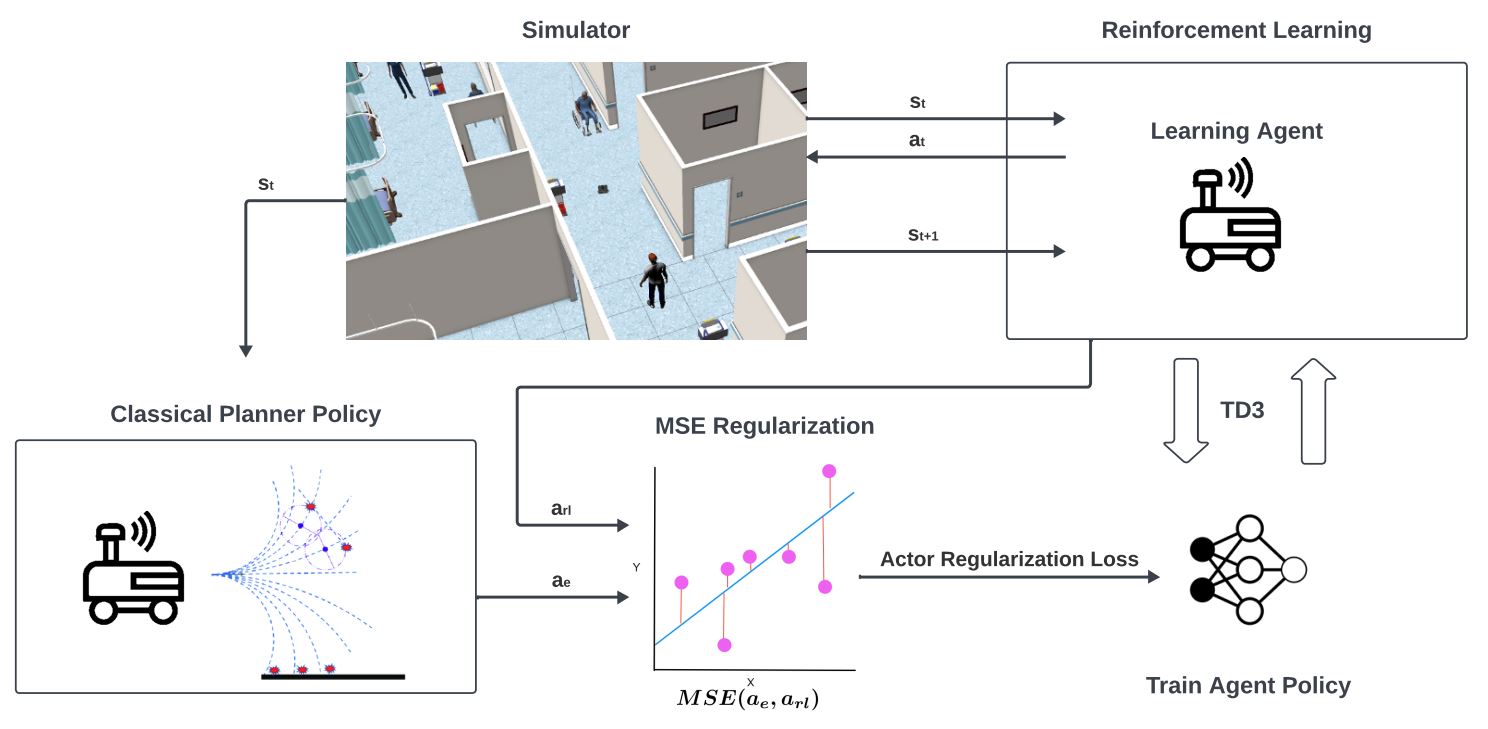}
    \caption{Framework of the learning strategy with a classical planner regularization.}
    \label{fig:csnrl}
\end{figure*}

\subsection{Reinforcement Learning with Demonstrations}

RL with demonstrations combines elements of IL into the RL procedure offering complementary strengths. IL enhances realism, and alleviates the challenge of reward design and extensive datasets, while RL enhances safety and robustness. 
IL can be used for the initialization of policies \cite{pfeiffer2018reinforced, liang2018cirl} and
expert demonstrations can be included in the experience replay buffer \cite{vecerik2017leveraging, nair2018overcoming} to sample both expert demonstrations and interactions. However, these methods can fail when randomness of exploration is needed at the beginning of training or when expert rewards are not accessible.

We are specifically interested in approaches that directly incorporate expert demonstrations during the training stage, explicitly integrating them into the learning process (e.g. \cite{kang2018policy, lu2022imitation, rlkl, dwarl}). This approach is especially useful when we seek a learned policy 
 similar to that of the expert.

Particularly relevant is the work of~\cite{rlkl},
 which uses an RL strategy that incorporates the KL divergence between imitative expert priors and the agent policy in the reward function. This approach regularizes the agent's behavior with the expert policy while  maintaining its exploration ability. 

We propose to exploit the same strategy but use classical planners as the "expert" prior policy. As explained before, classical planners for mobile robots are competent enough to be employed as baselines or, in this case, as expert priors. The added benefit of using available planners is that the collection of data-sets containing human-expert demonstrations is no longer required!

\subsection{Safe Planning}



Classical planners for mobile robots prioritize safety, and use well-understood and, therefore, trustworthy safety mechanisms,  but often behave sub-optimally. While RL algorithms often provide enhanced efficiency, 
complete safety assurance remains challenging, especially when using Deep RL. 

We focus on rule-based systems with fallback layers (e.g., \cite{safetynet, hybrid_control})
that assess ML planner outputs against predefined checks. If a plan is deemed unsafe, they can modify it, opt for alternative policies, or employ other strategies. These methods 
ensure predictable and reliable behavior through explicit rules. We employ a similar approach:
In critical scenarios, the policy is seamlessly switched from the RL policy to the classical planner. The switching module is trained to guarantee safety while minimizing transitions between the ML policy and the classical planner. This 
allows for the utilization of an optimal, human-compliant planning system when safety is assured while seamlessly transitioning to a practical system that prioritizes safety in critical settings.

Unlike \cite{whomtotrust}, which trains the RL algorithm and the switching module in unison,
We first train an RL agent, on top of which we train a fuzzy-control supervisor. This makes learning easier, prevents the RL policy from exploiting the fallback algorithm too much, and makes for more transparent behavior and hence, better safety guarantees.


\section{Background}

\subsection{Reinforcement Learning}

Reinforcement learning (RL) addresses the challenge of mastering the control of a dynamic system, characterized by a Markov decision process (MDP) denoted as \(M = (S, A, Tr, r, \gamma)\),
where \(S\) is the state space, \(A\) is the action space, $Tr:S\times A\rightarrow \Pi(S)$ is the transition function,  $r : S \times A \rightarrow \mathbb{R}$ is the reward function, and \(\gamma \in (0, 1]\) serves as a discount factor. The objective in RL is to develop a policy, described as a distribution over actions conditioned on states \(\pi(a_t | s_t)\), aiming to maximize the long-term discounted cumulative reward:

\begin{equation}
\max_{\pi} \mathbb{E}_{\tau \sim p_{\pi}(\tau)} \left[ \sum_{t=0}^{T} \gamma^t r(s_t, a_t) \right]
\end{equation}

Here, \(\tau\) denotes a trajectory, \(p_{\pi}(\tau)\) is the distribution of the trajectory under policy \(\pi\), and \(T\) represents the time horizon.

The Actor-Critic (AC) method is a prominent paradigm within RL, combining an actor (policy) and a critic (value function) for improved decision-making and evaluation:

\[
\text{Actor:\ \ } \pi(a_t | s_t; \theta^\pi)\quad
\text{Critic:\ \ } Q(s_t, a_t; \theta^Q)
\]

Here, \(\pi\) represents the policy function with parameters \(\theta^\pi\), and \(Q\) denotes the action-value function with parameters \(\theta^Q\). The actor aims to determine the optimal policy, while the critic evaluates the chosen actions.

\begin{figure}
    \centering
    \includegraphics[width=0.3\textwidth]{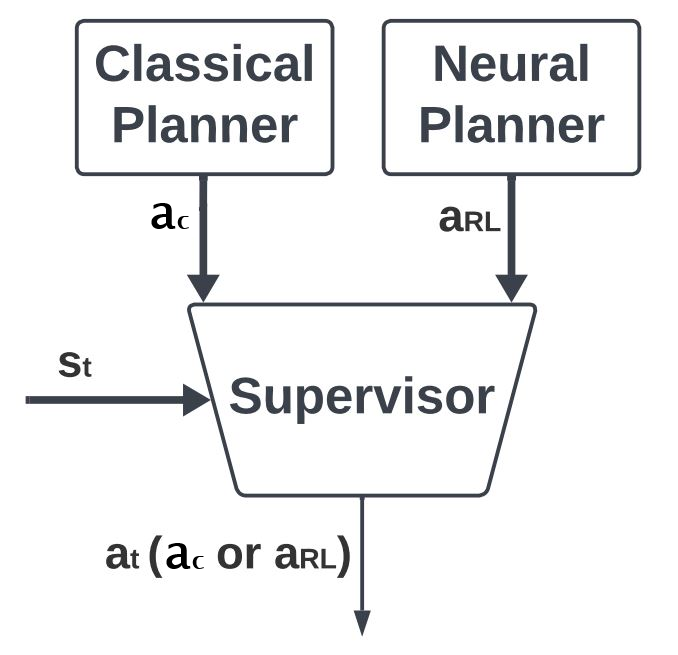}
    \caption{ A supervisor selects whether to switch from the RL to the Classical planner in critical situations.}
    \label{fig:supervisor}
\end{figure}

Drawing inspiration from \cite{rlkl}, we impose a constraint ensuring  similarity between the actor and the expert. This 
has two advantages: (1) A simplified exploration space concentrated around the expert policy, leading to decreased sampling complexity. (2) The policy  mimics the expert, reducing the need for a finely tuned reward function. To implement this, we introduce a regularization term in the actor loss update, defining the
\textit{regularized actor loss} as:

\begin{equation*}
\begin{split}
\mathcal{L}_{\text{actor}}(\theta^\pi) = \mathbb{E}_{s_t \sim \mathcal{D}} \bigg[ & -Q(s_t, \pi(s_t; \theta^\pi); \theta^Q) \\
& + \underline{\lambda \cdot \text{MSE}(\pi(s_t; \theta^\pi), \pi_{\text{expert}}(s_t))} \bigg]
\end{split}
\end{equation*}

Here, \(\mathcal{L}_{\text{actor}}(\theta^\pi)\) is the actor loss.
\(Q(s_t, a_t; \theta^Q)\) is the estimated action value by the critic.
\(\pi_{\text{expert}}(s_t)\) represents expert actions.
\(\lambda\) is the regularization strength.
\(\text{MSE}(\cdot, \cdot)\) is the mean squared error.

This formulation encourages the actor to generate actions that maximize expected rewards while regularizing them towards expert actions through the MSE term. This 
is particularly useful given sparse reward functions since the expert guides the learned policy. In our work, the "expert" is the classical planner. 
Unlike other approaches, such as \cite{dwarl}, which rely on a specific classical planner to guide the search, our method 
can seamlessly integrate with any expert.

\subsection{"Expert" Prior}
Our method assumes access to a good analytic algorithm for the task at hand. 
In robotics,
the ROS platform supplies multiple such algorithms.
In particular, for navigation, we employ
the \textit{move}{\textunderscore}\textit{base} framework alongside the classical local planner  DWA \cite{DWA}, 
from which we can derive actions at each time step that we use to
guide an AC algorithm.  The expert actions are remapped to an unused ROS \textit{topic}, while the RL actions are mapped to control the robot.

Unlike approaches relying on behavioral cloning, as seen in \cite{rlkl, imitationnotenough}, our method eliminates the need for a preceding learning step to imitate an expert. This makes our approach more cost-effective and less time-consuming.

\subsection{Safety Protocol}
Drawing inspiration from various hybrid navigation approaches such as \cite{hybrid_control, xiao2022motion, whomtotrust, safetynet}, we formulate  a practical protocol that can switch between a learning-based
planner 
and a classical planner. Following 
\cite{hybrid_control, safetynet}, we design this protocol with a focus on ensuring safety in critical scenarios,
recognizing that, in many instances, reliability and safety take precedence over swift navigation.

Similar to the approach of \cite{hybrid_control}, we employ deterministic rules
to determine when a switch to a safer navigation mode is imperative. However, we introduce a strategy to fine-tune these rules, aiming to optimize safety while minimizing the frequency of transitions to the safe planner.
More specifically, we utilize fuzzy rules and refine the membership functions through a genetic algorithm with the dual objectives of minimizing both transitions and critical situations.

\section{Architecture}\label{sec:Architecture}

The high-level architecture of the navigation algorithm is described in Fig.~\ref{fig:supervisor}: A supervisor module 
dynamically switches between a neural-net based policy and a classical planner, effectively maintaining a balanced 
overall policy 
that maintains safety, robustness and
efficiency in real-world scenarios.
We now describe its components in more detail. Section~\ref{sec:Methods} describes the methods used to train them.



\begin{figure}
    \centering
    \includegraphics[width=0.4\textwidth]{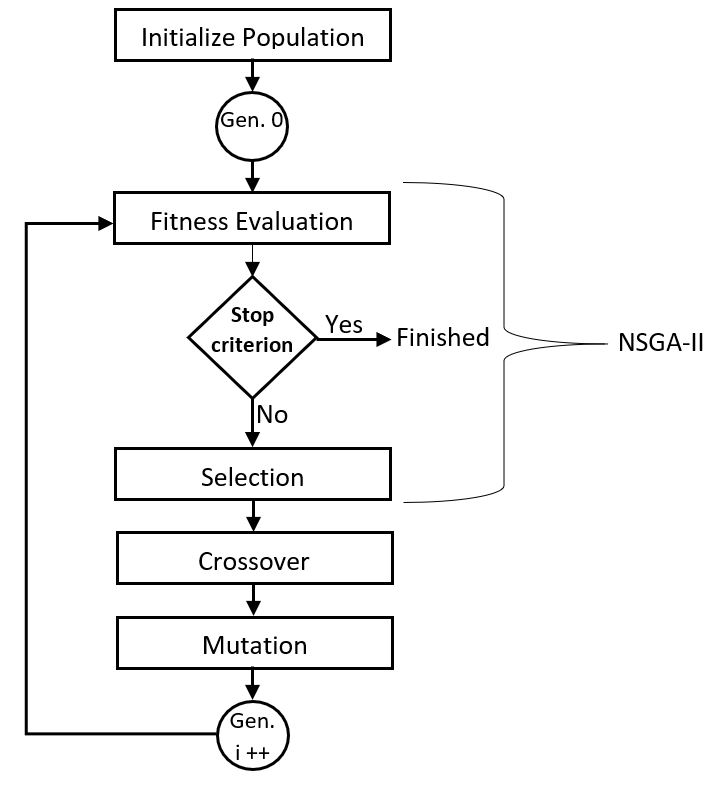}
    \caption{ Evolutionary Training Diagram for Supervisor. \textbf{Initialization:} Generate diverse populations with varying fuzzy membership values. \textbf{Fitness Evaluation:} Assess fitness through robot episodes, tracking critical situations and supervisor activations. \textbf{Selection:} Select values minimizing supervisor activations and critical situation occurrences.}
    \label{fig:nsga}
\end{figure}

\paragraph{Supervisor}

Generally, machine learning techniques perform well in navigation, but may fail badly in certain scenarios. 
To address this,  we adopt a  transparent and interpretable strategy that incorporates a rule-based safety mechanism directly during deployment. First, like many local planners, if the distance of the robot to the nearest obstacle $< 30cm$ then it moves backward. 
In addition, 
%
our \textit{Supervisor} module switches to a default safe policy
if the distance to the nearest obstacle is below a parameter called
{\em radius} below, which is determined using the following simple fuzzy rules: \\
\vspace{-6mm}

\begin{algorithm}
\caption{Fuzzy Rules for Supervisor Radius}
\begin{algorithmic}
\If{robot velocity is high}
    \State supervisor radius is big
\ElsIf{robot velocity is low}
    \State supervisor radius is small
\EndIf
\end{algorithmic}
\end{algorithm}

\paragraph{Neural Navigation Policy}
The neural network receives as input an egocentric 2D bird-eye view grayscale image of the local costmap (6x6 Meters). This image is created with a lidar sensor mounted on the robot. We also incorporate the current velocity of the robot and the location of the next waypoint (which is 2 meters away) as indicated by the global planner (a modified version of A*).
While its output is 
the linear and angular velocity command to the robot (continuous action space).

\begin{figure}
    \centering
    \includegraphics[width=0.45\textwidth]{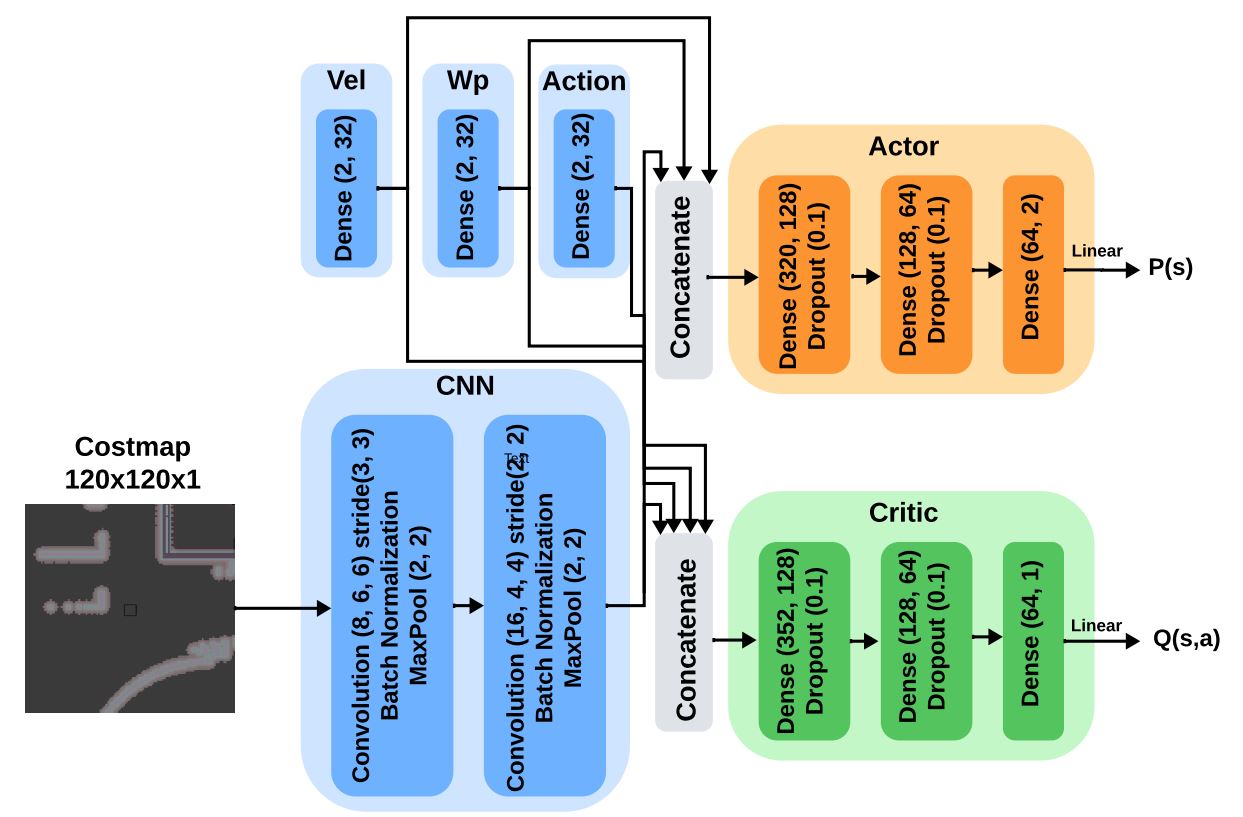}
    \caption{ Actor Critic neural networks. The Costmap is encoded using CNN's. The velocity (vel), waypoint, and action are concatenated.
    Two fully connected neural networks output the next action (actor) and the Q function (critic).}
   \vspace{-5mm}
    \label{fig:ac}
\end{figure}

We use a deep convolutional NN followed by fully connected layers for 
the actor and critic networks (See Fig.~\ref{fig:ac}).
We process the image and concatenate it with velocity and waypoint vectors and, in the case of the critic, the action.

\paragraph{Safe Policy}
The policy used when switching is the pure pursuit algorithm \cite{pure_pursuit} with a very low, fixed linear speed ($0.5 m/s$) towards the next waypoint, $30 cm$ away.
 This straightforward algorithm is  safe because of the following: (1) The $A^*$ variation used in this work creates a collision-free path given static obstacles. (2) The robot's low velocity affords ample time for individuals (dynamic obstacles) to adapt to its presence, and for $A^*$ to adapt to these obstacles.

\section{Methods}
\label{sec:Methods}
We train the neural policy and the genetic algorithm in simulation. We use a simulation platform of a realistic hospital \cite{aws} integrated with humans moving according to Social Forces, as described in \cite{Helbing1995SocialFM}. A
 computer with AMD® Ryzen 9 5900x12 CPU and GeForce RTX 4070 was used for simulation and training. We used Pytorch \cite{pytorch} for the implementation of the neural networks.

The robots used in simulation and real life are differential drive robots equipped with 2D Lidar sensors with a $360^0$ field of view. With this sensor, a 6x6-meter egocentric local costmap is generated. We use  \textit{move}{\textunderscore}\textit{base} to generate the global plan (which uses a variant of A*) and, from there extract intermediate waypoints, as well as the DWA, recommended actions.

Episodes involved the robot navigating between designated hospital rooms, with simulations sped up about five-fold, completing each episode in 8 to 14 seconds. The training approach, inspired by [26], separated Simulation, Robot, Task environments, and the Training algorithm.


\subsection{Neural Policy Training}
We use RL to develop a policy, training it for  $60,000$ episodes (around $6$ hours). The training process, illustrated in Fig. \ref{fig:csnrl}, refines the Twin Delayed DDPG (TD3) algorithm~\cite{TD3},
an online deep RL algorithm, through regularization from an expert, specifically the DWA planner.
%
TD3 uses
2 critics to mitigate overestimation bias and uses delayed updates to stabilize training by averting rapid policy changes and reducing divergence risk. It is known to provide improved sample efficiency compared to other AC algorithms.

Our modified algorithm, Expert-Enhanced TD3 (E2TD3),
uses  the
classical local planner DWA to modify two elements of TD3:
its replay buffer initialization and its policy update step. The pseudo-code is shown in Algorithm~\ref{algo:modified_TD3}. The changes are highlighted by underlining them, and are explained below. We note that similar modifications using an expert policy can be applied to
many other RL algorithms.  

\noindent{\bf Buffer Initialization.\ } TD3 uses a policy-gradient method to update the policy. The gradient is computed by sampling a mini-batch of experiences from the replay buffer.
E2TD3 initializes this buffer by recording experiences obtained by executing the expert (DWA) policy with some added normally distributed noise. This data pushes the algorithm in the general direction of the expert policy, while the use of an initially random policy and the noise added to the expert data provide some exploration bias. 

\noindent{\bf Regularization.\ } During learning, we augment new experiences 
$(s,a,r,s')$ with the expert recommended action $a_e$,
obtaining the quintuple $(s, a, \underline{a_e}, r, s')$.
In the \textit{Actor}'s learning stage, we add to the standard gradient update a regularizer in the form of
the MSE between the current policy and the expert policy:  $\lambda \cdot \text{MSE}(\pi_{\phi}(s), \pi_{\text{expert}}(s))$  (while we observed comparable results with alternative values, we opted for $\lambda = 1$). This 
has two advantages: (1) Simplified exploration space, reducing sampling complexity. (2) The actor mimics the expert policy, reducing the need for a finely tuned reward function.

\begin{algorithm}
\caption{The Expert-Enhanced TD3 (E2TD3) Algorithm}
\label{algo:modified_TD3}
\begin{algorithmic}
\State Initialize critic networks $Q_{\theta_1}$, $Q_{\theta_2}$, and actor network $\pi_{\phi}$ with random parameters $\theta_1, \theta_2, \phi$.
\State Initialize target networks: $\theta_{1}' \leftarrow \theta_1$, $\theta_{2}' \leftarrow \theta_2$, $\phi ' \leftarrow \phi$.
\State Initialize replay buffer $B$ 
\underline{by simulating the expert policy }
\underline{with Gaussian noise: $a=a_e+\epsilon,\ \epsilon \sim \mathcal{N}(0, \sigma)$.}

\For{$t = 1$ to $T$}
    \State Select action with exploration noise: $a \sim \pi_{\phi}(s) + \epsilon$, $\epsilon \sim \mathcal{N}(0, \sigma)$, and observe \underline{expert  action $a_e$}, reward $r$ and new state $s'$.
    \State Store transition tuple $(s, a, \underline{a_e}, r, s')$ in $B$.
    \State Sample mini-batch of $N$ transitions $(s, a, \underline{a_e}, r, s')$ from $B$.
    \State $\tilde{a} \leftarrow \pi_{\phi '}(s') + \epsilon$, $\epsilon \sim \text{clip}(\mathcal{N}(0, \tilde{\sigma}), -c, c)$.
    \State $y \leftarrow r + \gamma \min_{i=1,2} Q_{\theta_{i}'}(s', \tilde{a})$.
    \State Update critics: $\theta_i \leftarrow \text{argmin}_{\theta_i} \frac{1}{N} \sum_{i=1}^{N} (y - Q_{\theta_i}(s, a))^2$.
    \If{$t \mod d $}
        \State Update $\phi$ by the deterministic policy gradient:
        \[
        \begin{split}
        \nabla_{\phi} J(\phi) = & \frac{1}{N} \sum_{i=1}^{N} \nabla_a Q_{\theta_1}(s, a) \big|_{a=\pi_{\phi}(s)} \nabla_{\phi} \pi_{\phi}(s) \\
        & + \underline{\lambda \cdot \text{MSE}(\pi_{\phi}(s), \pi_{\text{expert}}(s))}
        \end{split}
        \]
        \State Update target networks:
        \[
        \begin{aligned}
        \theta_{i}' &\leftarrow \tau \theta_i + (1 - \tau) \theta_{i}' \\
        \phi ' &\leftarrow \tau \phi + (1 - \tau) \phi '
        \end{aligned}
        \]
    \EndIf
\EndFor
\end{algorithmic}
\end{algorithm}



\noindent{\bf Reward Function.\ } The  RL algorithm attempts to find a policy maximizing the expected sum of discounted rewards (Eq.~1) minus the regularizing term. 
The reward function is supposed to quantify desirable properties of the selected path. Our reward function  penalizes time and collisions, and rewards progress towards the waypoint. We tested the learning with two variations. In the first one, the reward function is more dense and greedy towards the waypoint: 
(1) $r(s,a) = r_{timestep} + r_{collision} + r_{progress\,waypoint}$\\
where $r_{timestep} = -0.5$ is a penalty for each time step, $r_{collision} = -vel - 1  \text{ \textit{if dist}} < 0.5 $ is a penalty for the robot being close and fast towards obstacles, and $r_{progress\,waypoint} = |dist(waypoint, postion_{t}) - dist(waypoint, postion_{t + 1})|$ is the distance traveled towards the current waypoint at each time step.

\begin{figure}
    \centering
    \includegraphics[width=0.45\textwidth]{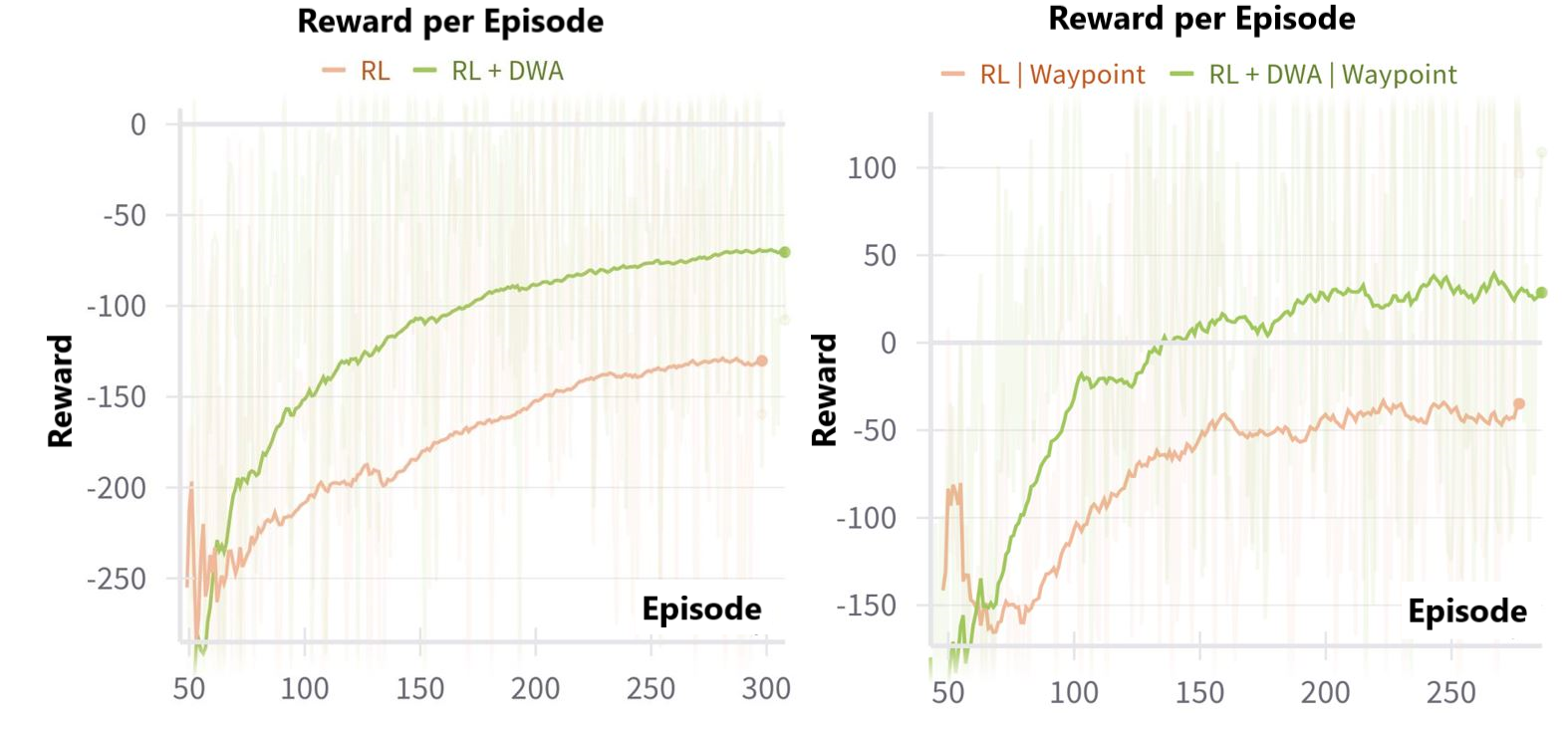}
    \caption{ Total reward per episode in simulations.
    We compare a plain RL policy with a regularized RL policy using the DWA classical planner, for two reward functions. As can be seen, RL + DWA outperforms the plain RL policy.}
    \label{fig:rewards}
\end{figure}

The second reward function is more sparse: the robot is rewarded only when it reaches the waypoint (with a margin of error):
(2) $r(s,a) = r_{timestep} + r_{collision} + r_{waypoint}$
where $r_{waypoint} = 10$
is a reward given when the robot reaches each specific waypoint.
The total reward during training can be seen in Fig. \ref{fig:rewards}. In both cases our approach shows better performance than plain RL. 

\subsection{Learning Supervisor Parameters}

The \textit{Supervisor} is a ruled-based mechanism that switches to a safe navigation policy as the robot nears obstacles, adjusting its radius dynamically based on velocity, increasing it with higher speeds. We define its parameters using
fuzzy rules that assign degrees of membership to variables for more flexible and nuanced decision-making compared to deterministic rules.

We optimize performance by fine-tuning membership function parameters of rule variables using a genetic algorithm. Variables include (1) \textit{Robot velocity} (low/high) and (2) \textit{Supervisor radius} (small/big). Membership functions for these variables are Gaussian, centered around $0m/s$, $1.5m/s$, and $0m$, $1.3m$, with variable standard deviations, learned using  the NonDominated Sorting Genetic Algorithm (NSGA-II), Fig.~\ref{fig:nsga}. We optimze two objectives: (1) Minimum switching. (2) Minimal critical scenarios. 

The Genetic Algorithm runs episodes of the robot moving in the hospital according to the learned RL policy while using the \textit{Supervisor} with a specific set of parameters for the fuzzy rules. For each episode, the fitness function contains the number of alternations and the number of critical scenarios. A critical scenario is defined as the robot being closer to obstacles than a fixed threshold (0.3 meters). 

\begin{figure}
    \centering
    \includegraphics[width=0.45\textwidth]{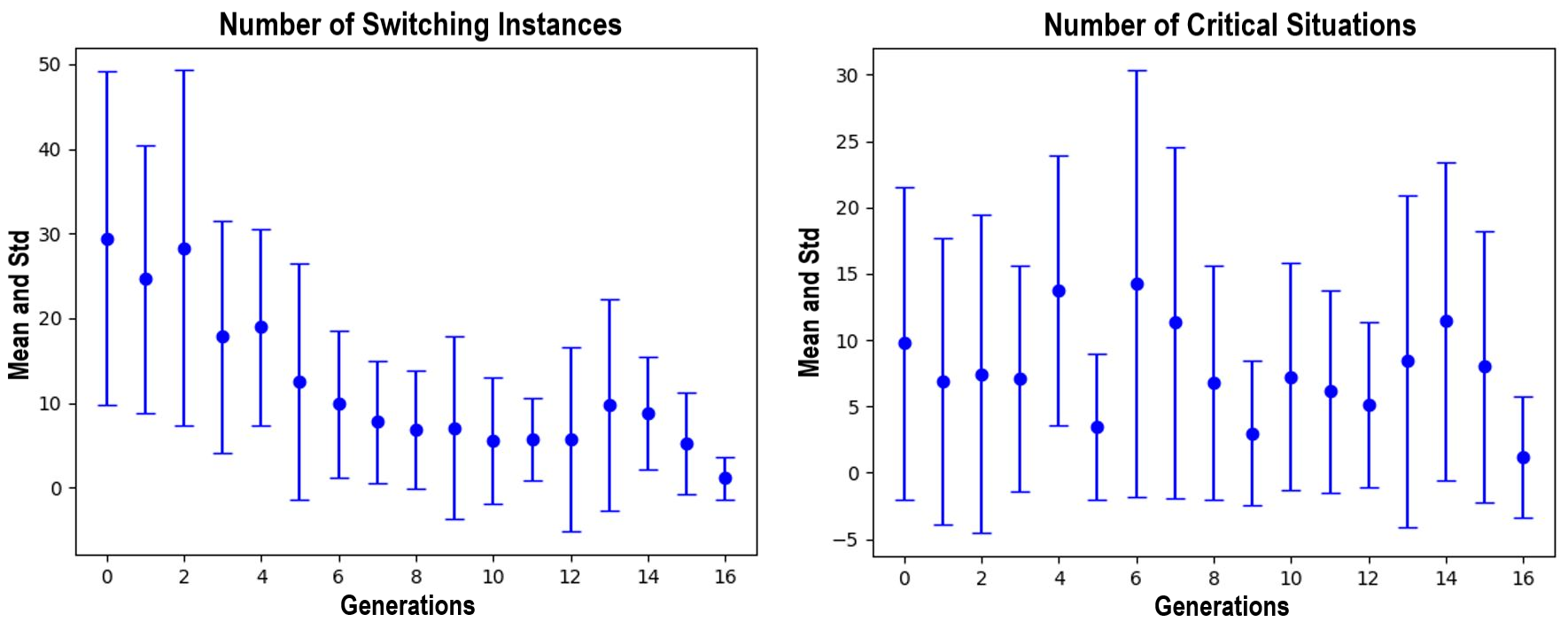}
    \caption{ Mean and STD of number of switching instances and critical situations across generations. We see that the Genetic Algorithm finds parameters that lower these values.}
    \label{fig:ga}
\end{figure}



\begin{table*}[!ht]
    \centering
    \caption{Performance statistics on $1e4$ steps ($\sim40$ episodes) per each algorithm in simulation. Average and $95\%$ confidence interval for: Total Reward, Total $r_{collision}$, Timesteps, MSE between DWA and algorithm actions (MSE(dwa,$\pi$)), and the number of critical situations (Critical) are presented. The algorithms tested are DWA, plain RL (RL), RL with DWA regularization (RL + DWA), and RL with regularization and supervisor module (RL + DWA + Supervisor).}
    \label{tab:table_simulation}
    
    \begin{tabularx}{\textwidth}{@{\extracolsep{\fill}} |c||X|X|X|X|}
         \hline
         \multicolumn{5}{|c|}{Performance Statistics \textbf{Simulation}. ((2) reward function)} \\
         \hline 
          & DWA & RL & RL + DWA & RL + DWA + Supervisor\\
          \hline 
         Total Reward       & $M=15.8 \pm 40.7$ & $M=-3.2 \pm45.8$ & $M=53.7 \pm37.2$ & $\boldsymbol{M=64.8 \pm30.9}$ \\
         \hline
         Total $r_{collision}$       & $M=-175.7 \pm 23.3$ & $M=-94.0 \pm18.3$ & $\boldsymbol{M=-83.5 \pm14.0}$ & $M=-89.5 \pm18.1$\\
         Timesteps       & $M=225.5 \pm 20.9$ & $M=296.3 \pm40.0$ & $M=226.1 \pm27.0$ & $\boldsymbol{M=221.9 \pm25.2}$\\
         MSE(dwa, $\pi$) & $M=0.0\%$ & $M=55.3\% \pm12.4\%$ & $M=23.2\% \pm3.0\%$ & $\boldsymbol{M=20.8\% \pm1.0\%}$\\
         Critical    & $M=4.8 \pm 3.3\%$ & $M=20.4\% \pm10.9\%$ & $M=11.4\% \pm6.2\%$ & $\boldsymbol{M=3.5\% \pm0.7\%}$\\
         \hline
     \end{tabularx}

\end{table*}

\begin{figure*}[t]
    \centering
    \begin{subfigure}[b]{0.24\textwidth}
        \centering
        \includegraphics[width=\textwidth]{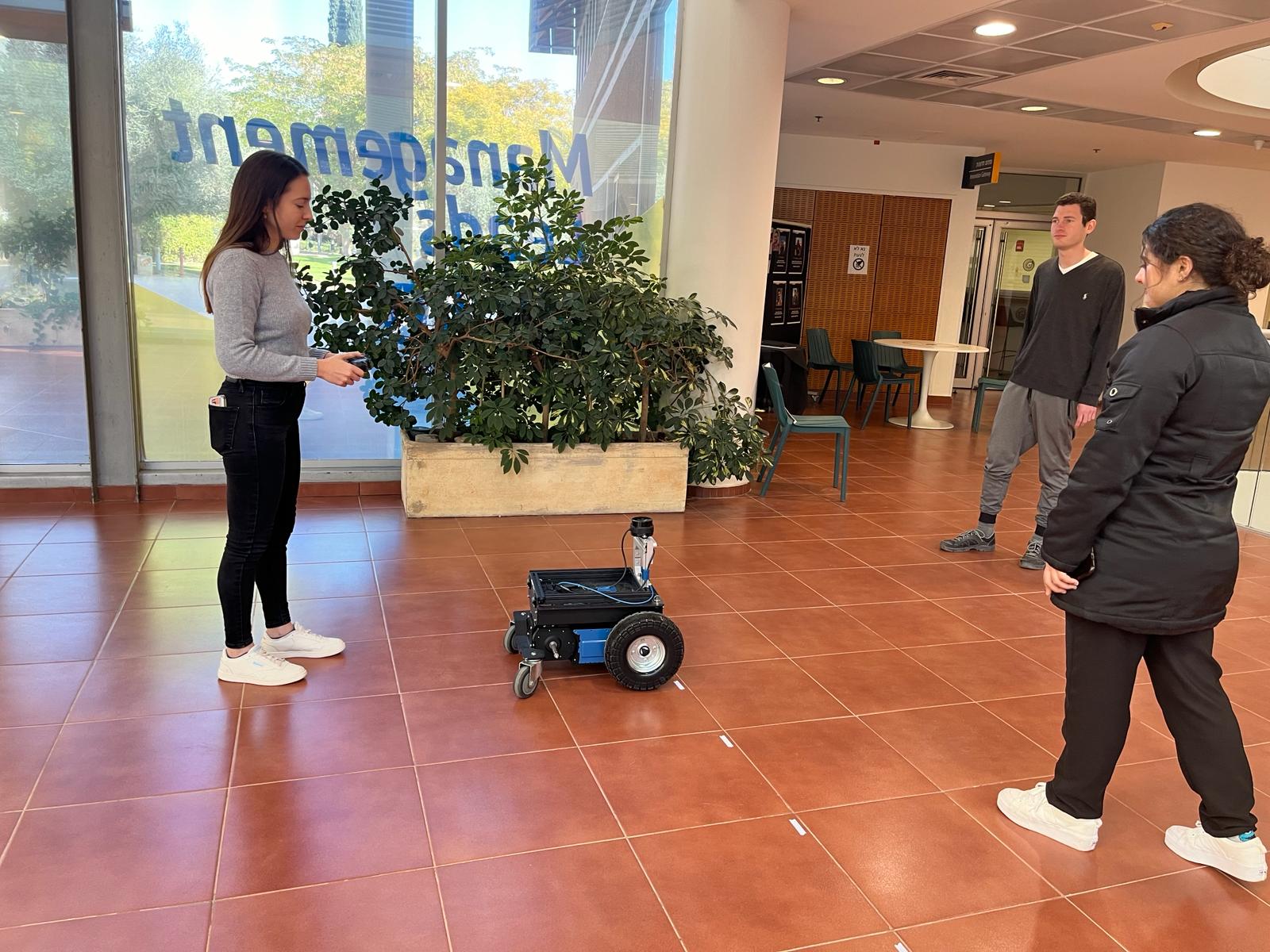}
        \label{fig:fig1}
    \end{subfigure}
    \begin{subfigure}[b]{0.24\textwidth}
        \centering
        \includegraphics[width=\textwidth]{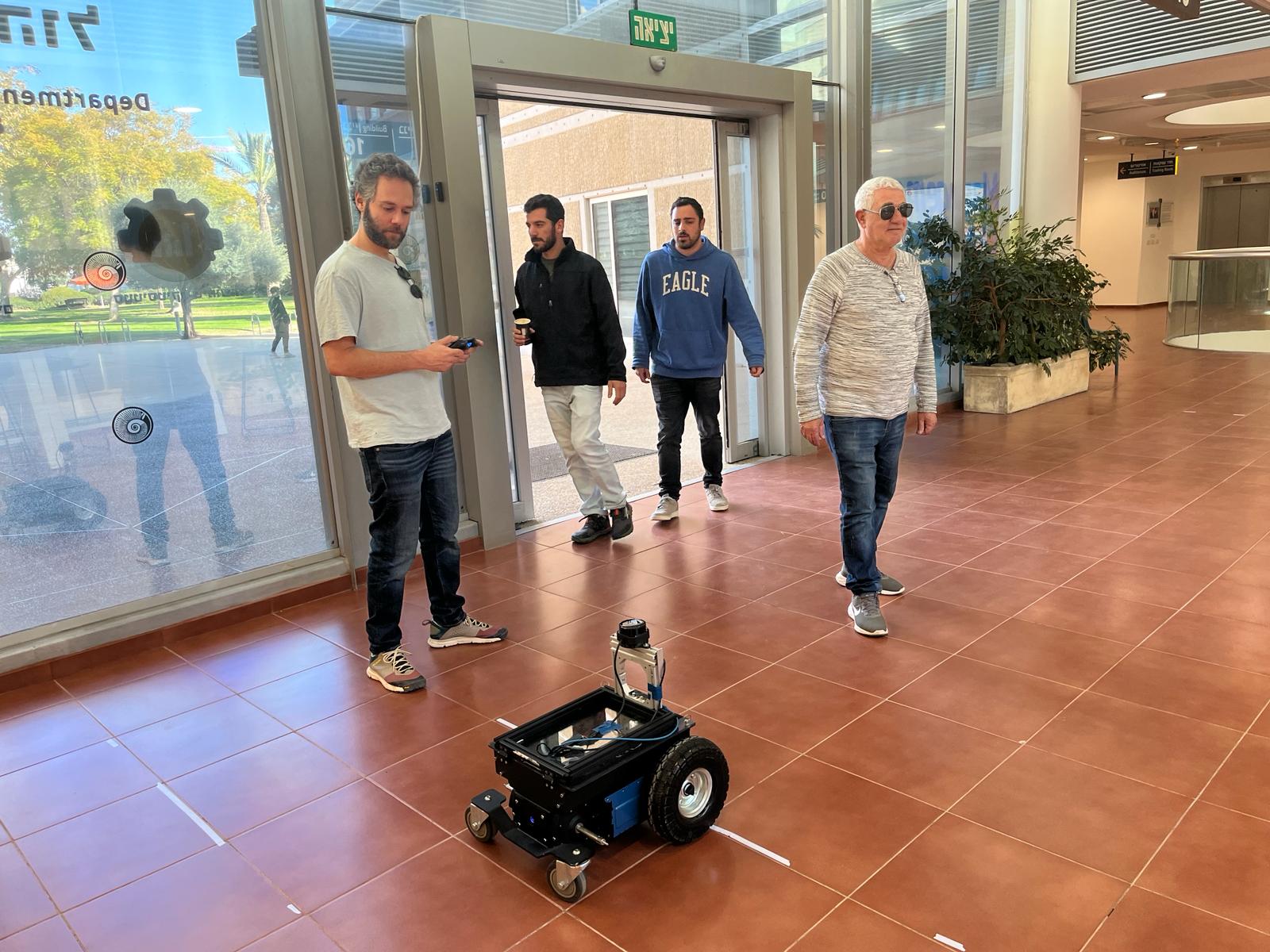}
        \label{fig:fig2}
    \end{subfigure}
    \begin{subfigure}[b]{0.24\textwidth}
        \centering
        \includegraphics[width=\textwidth]{figures/4_2.jpg}
        \label{fig:fig3}
    \end{subfigure}
    \begin{subfigure}[b]{0.24\textwidth}
       \centering
        \includegraphics[width=\textwidth]{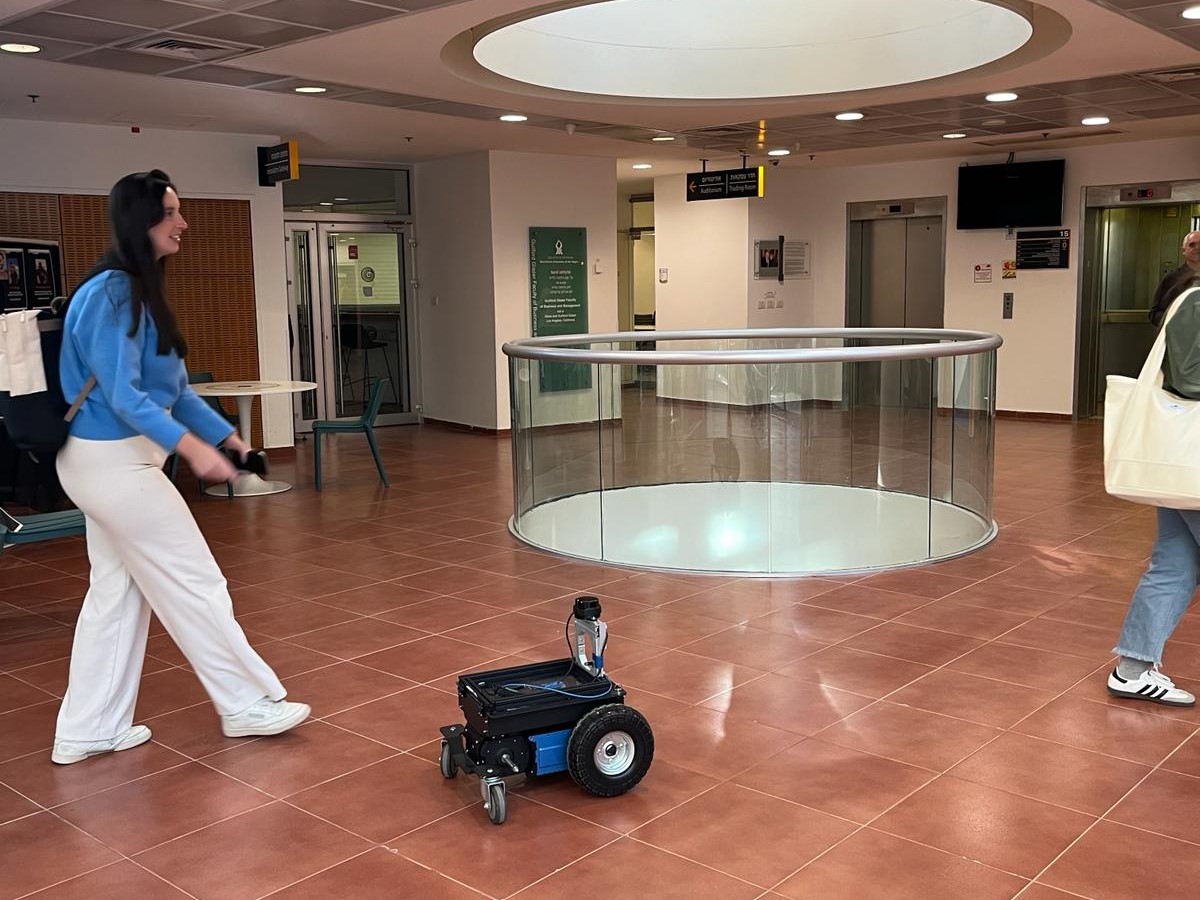}
        \label{fig:fig4}
    \end{subfigure}
    \caption{Rover Zero 3 robot navigating autonomously in different environments.}
    \label{fig:multiple_figures}
\end{figure*}

\begin{table}[!ht]
    \centering
    \caption{Performance statistics in $18$ real test cases 
    }
    \label{tab:table_physical}
    
    \begin{tabularx}{\columnwidth}{|c||X|X|}
         \hline
         \multicolumn{3}{|c|}{Performance Statistics \textbf{Real-life}} \\
         \hline 
          & DWA & RL + DWA + Superv.\\
          \hline 
         Time       & $\boldsymbol{M=33.3 \pm 7.4}$ & $M=36.0 \pm10.2$ \\
         \hline
         Negative Score       & $M=13.6\% \pm 3.7\%$ & $\boldsymbol{M=4.7\% \pm2.2\%}$\\
         \hline
         Interventions      & $M=1.3 \pm 0.4$ & $\boldsymbol{M=0.2 \pm0.3\%}$\\
         \hline
     \end{tabularx}
\end{table}







The supervisor module was trained for $16$ generations, each generation with $16$ individuals. Individuals were episodes of the robot moving through hospital rooms with the learned RL + DWA policy, augmented with randomized actions. The robot would switch to a low-velocity pure pursuit algorithm as a function of its velocity and proximity to obstacles, determined by the current values of fuzzy rules. The fitness function was evaluated based on the number of switching instances and critical situations (extreme closeness to obstacles) in an episode.
As can be seen in Fig. \ref{fig:ga}, NSGA-II learned to reduce the number of switching instances while also reducing the number of critical situations.

\section{Experiments}

In our study,  we aim to show the efficacy of our algorithm, Regularized RL with a Supervisor, by demonstrating that it can:  (1) be easily implemented with an accessible expert policy (2) improve the expert policy while maintaining proximity to it (3) learn faster, and possibly reach better cumulative rewards than plain RL (4) effectively integrate a supervisor module to diminish unexpected critical situations, thereby rendering it suitable for practical applications.

\subsection{Simulated Experiments}

We first test and compare DWA, RL, RL + DWA (regularized RL), and RL + DWA + Supervisor (regularized RL with a supervisor) in a simulated realistic hospital with dynamic people moving according to social forces, see Table \ref{tab:table_simulation}.  Each algorithm was run for $1e4$ timesteps (around $40$ episodes). The metrics for comparison were: Total Reward, Total $r_{collision}$, Timesteps, MSE between DWA and algorithm actions (MSE(dwa,$\pi$)), and the number of critical situations (Critical). A bigger Total Reward and Total $r_{collision}$ are better. Smaller Timesteps, MSE(dwa,$\pi$), and critical situations are preferred.

RL learns a decent navigation policy  capable of reaching distinct rooms in the hospital. However, it can get stuck in local minima in complex situations (narrow passages or small rooms). The expert (DWA) is better than it in most metrics.

RL + DWA already shows an improvement (Total Reward, $r_{collision}$) or similar performance (Timesteps) both over RL and DWA. Furthermore, RL + DWA stays closer to DWA ($MSE=23.2\%$) than RL ($MSE=55.3\%$). However, RL + DWA has a high percentage of critical situations ($11.4\%$), which makes it unreliable.

RL + DWA + Supervisor has the best or similar performance in all metrics. Most importantly, it lowers the number of critical situations ($3.5\%$) to a lower threshold than RL + DWA ($11.4\%$) and even of DWA ($4.5\%$). 

This shows the advantage of guiding the search space of machine learning algorithms when a proficient yet suboptimal expert is available. Our algorithm can be straightforwardly implemented in other AC implementations whether in online or offline RL, with different experts. The inclusion of the Supervisor module enhances its practicality and reliability. This is important in real-life applications.

\subsection{Physical Experiments}
We performed physical experiments using a Roverrobotics Rover Zero robot wth a SLAMTEC RPLidar S1 and an Intel RealSense T265 Tracking Camera. 
The robot's computer (Intel NUCi7) executed the algorithms, and external control of the robot was established through the Master-Slave ROS protocol from a separate computer. A  Dualshock 3 joystick was connected remotely to the robot.

We tested our algorithm, RL + DWA + Supervisor, using the policy learned in simulation with no additional fine-tuning, against DWA in various university areas (corridors, corners, entrances, hall) with static obstacles and people moving around the robot. The robot navigated autonomously using the \textit{move}{\textunderscore}\textit{base} plugin to generate a global plan with waypoints and  DWA or our algorithm to control its actions. At each moment, a person could press the joystick's $X$ button to indicate dissatisfaction and could also control the robot's movement in critical situations ({\em Intervention}). 

We carried $18$ trials, 9 for each algorithm, with $3$ different people recording Time, Negative Score, and Interventions. The results are shown in Table~\ref{tab:table_physical}. Negative Score is the percentage of time of dissatisfaction with the robot's performance reflecting suboptimal driving situations like proximity to obstacles, slow movement in open areas, or significant deviation from the goal.

While DWA exhibited quicker completion times, our algorithm obtained a lower negative score and fewer interventions. It demonstrated faster performance in open areas, efficiently navigated in clustered spaces, maintained a greater distance from people, passed through doors more effectively, and avoided collisions thanks to the Supervisor module.  However, it tends to oscillate more due to policy switching.

\section{Conclusion}

This work introduces a framework that uses a well-known, trustworthy algorithm  to help regularize the actor within an AC-based deep RL algorithm for the autonomous navigation of mobile robots. 
It then learns fuzzy rules for optimizing a safety module that switches between the learned policy and a safe but suboptimal one in critical situations. This pragmatic approach is able to improve
a classical algorithms through reinforcement learning while preserving practicality and safety. Unlike most imitation learning algorithms, it requires no expert human input. 

Our algorithm was able to enhance a prior algorithm (DWA, here) with low training time, adopting some of the key features of DWA while introducing new elements to address specific scenarios demonstrating superior performance on key measures. Additionally, the inclusion of the Supervisor module helped ensure safety at all times, making it a practical and reliable algorithm. 

\bibliographystyle{IEEEtran}
\bibliography{IEEEabrv,mybibfile}

\end{document}